%% file: neurips_2026.tex
\documentclass{article}

\PassOptionsToPackage{numbers,sort&compress}{natbib}
 \usepackage[preprint]{neurips_2026}

\usepackage[utf8]{inputenc} 
\usepackage[T1]{fontenc}    
\usepackage{hyperref}       
\usepackage{url}            
\usepackage{booktabs}       
\usepackage{amsfonts}       
\usepackage{nicefrac}       
\usepackage{microtype}      
\usepackage{xcolor}         
\usepackage{graphicx} 
\usepackage{makecell}
\usepackage{multirow}
\usepackage{colortbl}
\usepackage{caption}
\usepackage{amsmath}
\usepackage{subcaption}
\usepackage{amssymb}
\definecolor{best_color}{rgb}{1, 0.7, 0.7}
\definecolor{second_color}{rgb}{1, 0.85, 0.7}
\definecolor{third_color}{rgb}{1, 1, 0.7}
\newcommand{\best}{\cellcolor{best_color}}
\newcommand{\second}{\cellcolor{second_color}}
\newcommand{\third}{\cellcolor{third_color}}

\title{Seg3DParts: Segmentation-Grounded Controllable Part-Level 3D Generation}

\author{%
  \textbf{Jiantao Lin}$^{1,*}$ \quad
  \textbf{Meixi Chen}$^{1,*}$ \quad
  \textbf{Yingjie Xu}$^{1,3,*}$ \quad
  \textbf{Chenbo Fu}$^{1}$ \quad
  \textbf{Leyi Wu}$^{1}$ \\
  \textbf{Hao Chen}$^{2}$ \quad
  \textbf{Yinchuan Li}$^{3}$ \quad
  \textbf{Ying-Cong Chen}$^{1,2,\dagger}$ \\[2pt]
  {\normalfont $^{1}$The Hong Kong University of Science and Technology (Guangzhou)} \\
  {\normalfont $^{2}$The Hong Kong University of Science and Technology} \quad
  {\normalfont $^{3}$Knowin AI} \\[2pt]
  {\normalfont $^{*}$Equal contribution. \quad $^{\dagger}$Corresponding author.}
}

\begin{document}

\maketitle

\input{sec/0_abstract}

\input{sec/1_introduction}

\input{sec/2_related_work}

\input{sec/3_method}

\input{sec/5_experiments}

\input{sec/6_conclusion}

\bibliographystyle{IEEEtran}

\bibliography{ref.bib}

\newpage
\appendix

\input{sec/7_appendix}



\end{document}

%% file: sec/0_abstract.tex
\vspace{-3em}
\begin{figure}[h]
    \centering  \includegraphics[width=0.9\textwidth]{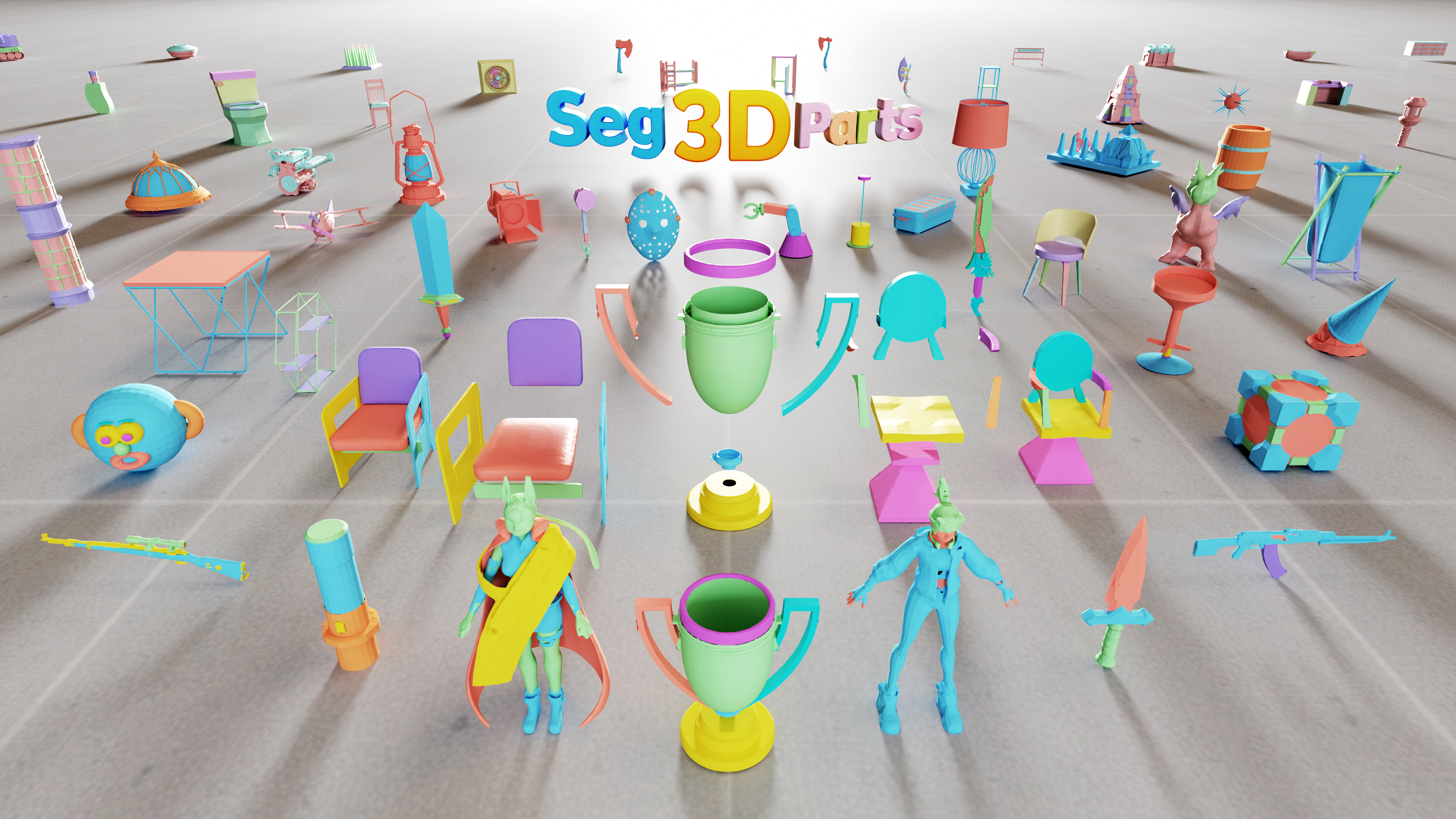}
  \centering
  \label{fig:teaser_figure}
\end{figure}

\begin{abstract}
Part-level 3D assets are essential for editing, reassembly, and interaction, yet recovering such structure from a single image remains challenging due to occlusion, ambiguous boundaries, and the need for coherent multi-part reasoning.
Existing approaches struggle to achieve both controllable part-level generation and coherent multi-part structure, as part identity and spatial allocation are typically inferred implicitly. We present Seg3DParts, a segmentation-grounded framework for controllable part-level 3D generation from a single image.
By treating segmentation as an explicit grounding signal, our method defines part identity during generation, enabling each component to be anchored to a corresponding image region.
To ensure coherent assemblies, we introduce structured cross-part interaction that allows components to exchange global context throughout the generative process. 
As a result, Seg3DParts directly generates well-aligned part meshes in a shared canonical space without post-hoc alignment, supporting flexible and controllable decomposition.
We further introduce PartObjectNet, a large-scale dataset with over 200K objects and 1M annotated parts.
Experiments demonstrate that Seg3DParts achieves superior geometry quality, cross-part coherence, and part-level controllability over existing methods.

\end{abstract}

%% file: sec/1_introduction.tex
\section{Introduction}
Generating part-level 3D structure from a single RGB image is fundamentally ill-posed.
Local appearance alone is often insufficient to determine precise part boundaries,
especially when adjacent components share similar materials or exhibit weak shading cues.
Occlusion further complicates the problem, as important functional components may be
partially or entirely invisible in the input view, providing little or no direct
2D evidence for their geometry.
Beyond individual parts, a model must also reason about cross-part spatial relationships
in 3D, ensuring that components attach correctly and maintain plausible relative scale
and placement.
A practical system therefore needs to recover missing geometry, localize parts reliably,
and reason coherently across multiple interacting components under severe single-view ambiguity.

\begin{figure}[t]
    \centering  \includegraphics[width=0.9\textwidth, trim=0 18 0 5,clip]{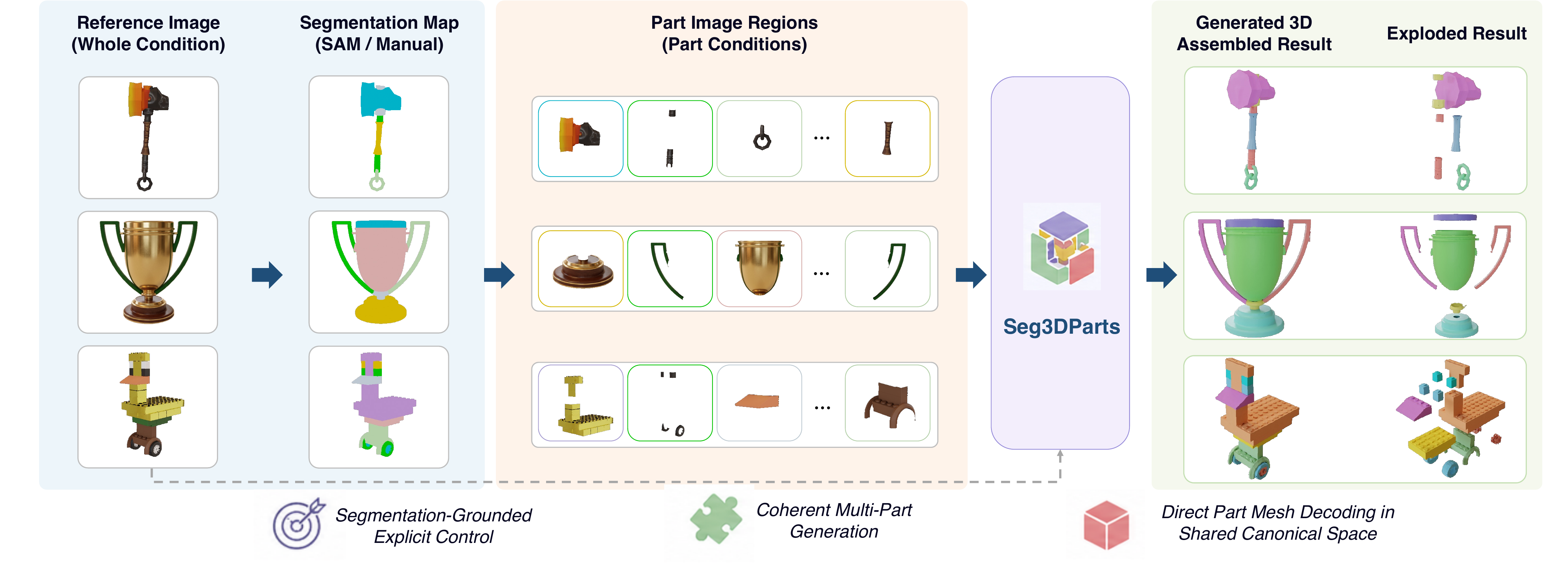}
  \centering
  \caption{Conditioned on reference image and per-part image regions derived from a segmentation map , \textbf{Seg3DParts} jointly generates all part meshes in a shared canonical space, yielding controllable decomposition and coherent multi-part assemblies.}
  \vspace{-1.em}
  \label{fig:teaser}
\end{figure}

Existing approaches to part-level 3D generation generally follow two paradigms. Decomposition-based pipelines explicitly segment an object into parts and reconstruct each component independently by completing missing geometry before assembling them into a full shape~\cite{holopart, xpart, autopartgen,partgen,SAMPart3D,liu2025partfield}. While this design provides direct control over part decomposition, reconstruction is driven primarily by local geometric cues within each segment, with limited modeling of cross-part relationships.
As a result, these methods often produce inconsistent scale, misalignment, or implausible assemblies, especially when parts are heavily occluded or contain large missing regions.
In contrast, joint part-structured generative models represent multiple components within a unified latent space and generate them in an end-to-end manner~\cite{partcrafter, copart,partpacker,fullpart}.
By jointly modeling all components, these methods encourage global coherence across parts.
However, parts are encoded as implicit latent slots without explicit semantic grounding or spatial anchoring.
Consequently, part identity, correspondence, and placement must be inferred during generation, which can lead to ambiguous decomposition and limited controllability.

Despite their differences, both paradigms share a common limitation: part identity and spatial allocation are treated as implicit variables that must be inferred rather than explicitly specified.
This implicit formulation makes it difficult to achieve both precise part-level control and coherent multi-part generation, particularly under occlusion or limited visual evidence.

We instead introduce a formulation where part identity is explicitly specified and controllable via segmentation, rather than implicitly inferred during generation.
Specifically, we reformulate part-level 3D generation as a segmentation-grounded conditional generation problem, where segmentation defines part identity and anchors each component to its corresponding image region. Under this formulation, Seg3DParts enables controllable and coherent multi-part 3D generation from a single image (Fig.~\ref{fig:teaser}).

At its core, Seg3DParts conditions each component on localized appearance cues derived from segmentation, allowing part-specific geometry to be generated from its corresponding image region.
This explicit grounding provides direct control over part identity and improves robustness under occlusion and ambiguous boundaries.

To ensure coherent multi-part structure, Seg3DParts further introduces a part-level interaction mechanism that enables components to exchange global structural context throughout the generative process.
This structured interaction allows each part to reason about the shape, scale, and placement of others while preserving explicit part identities.

As a result, Seg3DParts learns a coherent multi-part representation from which independent part meshes are directly decoded in a shared canonical space, eliminating post-hoc alignment and enabling flexible, segmentation-driven, controllable generation. 
We also construct PartObjectNet, a large-scale dataset of part-separated 3D objects with rich part-level annotations.
Together, these design principles shift part-level 3D generation from implicit allocation to explicitly grounded conditional generation.

Our contributions are summarized as follows:
\begin{itemize}
    \item \textbf{A segmentation-grounded formulation for part-level 3D generation.}
    We introduce a formulation that explicitly anchors part identity to segmentation regions, enabling controllable and spatially grounded generation.

    \item \textbf{A unified framework for controllable and coherent multi-part generation.}
    We instantiate this formulation with segmentation-grounded conditioning and structured cross-part interaction, enabling joint reasoning over shape, placement, and inter-part relationships.

    \item \textbf{A large-scale dataset of part-separated 3D objects.}
    We construct \textit{PartObjectNet}, a dataset containing over 200K objects with more than 1M annotated parts, providing rich and structured supervision for learning complete and coherent part-level geometry.
\end{itemize}

%% file: sec/2_related_work.tex
\section{Related Work}
\subsection{3D Generation}
Recent advances in 3D generation have enabled the synthesis of high-quality 3D objects from text, images, or learned shape distributions~\cite{lin2025kiss3dgen, zhang2024clay, xiang2025structured, zhao2025hunyuan3d, zero123, wonder3d, jia2025ultrashape, wu2025direct3d,yang2025pandora3d,li2025triposg,wu2024direct3d,zhang20233dshape2vecset}.
Early approaches often relied on category-specific models or 2D-driven pipelines, while more recent methods adopt 3D-native generative models operating in geometry-aware latent spaces, such as 3D latent diffusion frameworks~\cite{li2025sparc3d, zhang2024clay, xiang2025structured}.
These models significantly improve geometric fidelity and scalability, and some recent works further enable direct mesh generation with well-structured topology and clean connectivity, without post-processing~\cite{siddiqui2024meshgpt, dai2025meshcoder}.
Despite this progress, most existing 3D generation methods focus on whole-object synthesis and produce monolithic representations without explicit part structure.
While effective for visualization and rendering, such representations offer limited support for fine-grained editing, reassembly, or part-level control, motivating growing interest in part-level 3D generation.

\subsection{Part-Level 3D Generation}
Recent work on part-level 3D generation moves beyond monolithic object synthesis toward explicit component-based modeling.
Existing methods can be broadly categorized into two paradigms: decomposition-based pipelines, which explicitly segment a whole object into surface parts and process each component separately; and part-structured generative models, which represent multiple components jointly within a unified generative framework.

\subsubsection{Decomposition-Based Part Generation Pipelines}  
Decomposition-based pipelines approach part-level 3D generation by explicitly dividing a complete object into surface parts and reconstructing each component through part-level geometry completion.
Given an whole object mesh, these methods first perform surface segmentation and then recover missing geometry for individual parts before assembling them into a full shape~\cite{holopart, xpart, liu2025partfield, SAMPart3D, GeoSAM2}.
Representative work such as HoloPart~\cite{holopart} exemplifies this paradigm by performing holistic surface segmentation followed by part-wise geometry completion.
While explicit decomposition enables controllable manipulation of individual components, part reconstruction is primarily guided by local surface cues within each segment.
Cross-part consistency is enforced only after each part has been completed, rather than influencing how parts are reconstructed in the first place.
As a result, these methods struggle when parts are heavily occluded or contain large missing surface regions, where local geometry alone provides insufficient constraints to infer complete and semantically consistent components.
This often leads to misaligned assemblies or inconsistent relative scales in the reconstructed shapes.

\subsubsection{Part-Structured Generative Models}
Part-structured generative models synthesize objects by jointly representing multiple components within a unified latent space and generating them in an end-to-end manner~\cite{partcrafter, partpacker, moca, autopartgen, fullpart}.
By coupling parts during generation, these methods encourage global structural coherence and produce plausible multi-part assemblies.

Despite these advantages, parts are typically encoded as implicit latent slots without explicit semantic specification.
As a result, there is no stable correspondence between latent slots and interpretable object components, making it difficult to explicitly assign, constrain, or manipulate individual parts.
The number, identity, and granularity of parts are therefore often implicitly fixed by model design, limiting flexibility under alternative decompositions.

Recent work such as OmniPart~\cite{ominipart} explores incorporating segmentation into part-level generation by partitioning a voxel representation using global segmentation (e.g., SAM).
However, segmentation is used as a global partitioning signal rather than explicitly defining part identities during generation.
As a result, part allocation and spatial assignment remain implicitly determined, which can lead to ambiguity in part boundaries and merged components, especially under occlusion.

Together, these methods rely on implicit modeling of part identity and spatial allocation, rather than explicitly grounding parts to observable signals.

%% file: sec/3_method.tex
\section{Method}
\begin{figure*}[t]
    \centering
    \includegraphics[width=\linewidth]{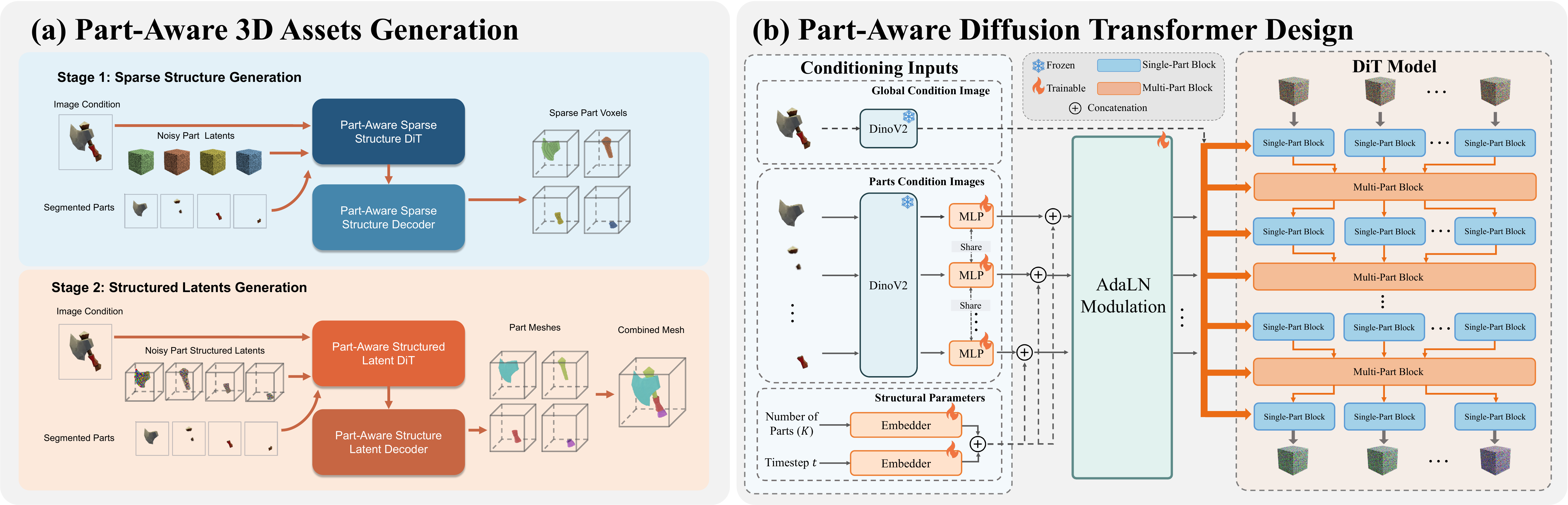}
    \vspace{-1em}
    \caption{
    \textbf{Overview of Seg3DParts.}
    \textbf{(a) Part-aware 3D asset generation pipeline.}
    Seg3DParts adopts a two-stage framework.
    Stage~1 predicts part-aware sparse voxel structures from the input image and segmentation.
    Stage~2 generates geometry-aware latents for each part voxel, which are decoded into meshes aligned in a shared canonical space for direct assembly.
    \textbf{(b) Part-aware DiT.}
    The DiT used in both stages follows the same part-aware design.
    Global image features from frozen DINOv2~\cite{dinov2} are injected via cross-attention, while part-specific features and structural parameters modulate the network through AdaLN~\cite{adaln}.
    The DiT alternates single-part blocks for intra-part modeling and multi-part blocks for cross-part interaction, enabling coherent multi-part generation.
    }
    \vspace{-1.5em}
    \label{fig:pipeline}
\end{figure*}

We formulate part-level 3D generation as a segmentation-grounded conditional generation problem, where segmentation explicitly defines part identity and provides a direct interface for controllable generation.
To this end, we propose \textbf{Seg3DParts}, a segmentation-grounded multi-part 3D generation framework built upon the structured latent representation and rectified-flow pipeline of TRELLIS~\cite{xiang2025structured}.
As illustrated in Fig.~\ref{fig:pipeline} (a), given an input image and part segmentation, Seg3DParts conditions each component on localized appearance cues for part-specific geometry generation.
We first introduce the structured latent backbone based on TRELLIS, followed by segmentation-grounded part conditioning and multi-part latent interaction for coherent generation. Finally, we describe canonical-space decoding and training objectives.

\subsection{Preliminaries}
\label{sec:Preliminaries}
Seg3DParts builds upon the structured latent 3D generation framework of TRELLIS, which represents 3D shapes using sparse voxel structures and synthesizes geometry through a two-stage generative process.

TRELLIS consists of \emph{Sparse Structure Generation} and \emph{Structured Latents Generation}.
In the first stage, a VAE together with a rectified-flow DiT models sparse voxel occupancy, defining the global spatial structure without encoding geometric details.
In the second stage, conditioned on the sparse structure, a separate VAE and DiT generate structured latents that capture fine-grained geometry and are decoded into surface meshes.
Following TRELLIS, we refer to these components as the \emph{sparse structure VAE/DiT} and \emph{structure latent VAE/DiT}, respectively.

In this work, we adopt the sparse voxel representation and the two-stage VAE-based rectified-flow framework of TRELLIS as our backbone.
Unless otherwise specified, all VAEs and DiTs described in the following sections are instantiated in a part-aware manner, where latent representations are maintained per part rather than for the whole object.

\subsection{Segmentation-Grounded Part Conditioning}
\label{sec:Segmentation}

Seg3DParts grounds each generated component in localized image evidence via segmentation-conditioned features, enabling explicit part identity specification during generation, as illustrated in Fig.~\ref{fig:pipeline} (b). 
This design enables explicit part-level specification and provides direct control over decomposition and generation from a single input view, and is applied to both \emph{part-aware sparse structure DiT} and \emph{part-aware structure latent DiT}.

Given an input RGB image $I$, for each part $k$, we obtain a part segmentation either from an external segmentation model (e.g., SAM~\cite{sam}) or from user-provided annotations. We extract the corresponding region from $I$ to form a part-conditioning image $I^{k}_{\text{part}}$, which provides localized visual evidence for that component.

Each part-conditioning image $I^{k}_{\text{part}}$ is encoded by a frozen DINOv2 backbone to extract high-level visual features, followed by a lightweight trainable MLP that maps these features into a compact appearance embedding $\mathbf{v}_k$.
The embedding captures the visible attributes of the corresponding part and remains robust to partial or missing observations caused by occlusion.
We concatenate $\mathbf{v}_k$ with the diffusion timestep embedding and an encoding of the total number of parts $K$ to form a part-specific conditioning vector $\mathbf{c}_k$.
This vector is projected to the modulation dimension and injected into all DiT layers via Adaptive Layer Normalization (AdaLN).

In addition to part-wise conditioning, the global-conditioning image $I$ is encoded by DINOv2 to provide global contextual information, which is injected to the model via cross-attention.

By separating part-level modulation from global contextual conditioning, Seg3DParts ensures that each component is grounded in its corresponding image region while remaining consistent with the overall object structure.
This design enables fine-grained control over part identity and appearance, supports varying decomposition granularities, and allows the model to handle occluded or invisible components.

\subsection{Multi-Part Latent Interaction}
\label{sec:Interaction}

While segmentation-grounded part conditioning specifies the appearance of individual components,
coherent 3D assemblies further require explicit reasoning across parts during generation.
Relative placement, scale consistency, and structural compatibility cannot be reliably inferred
from per-part cues alone, especially under occlusion or limited visual evidence.

Seg3DParts introduces \emph{multi-part latent interaction} throughout the generative pipeline.
All VAEs and rectified-flow DiTs operate on part-separated latent representations and
periodically enable information exchange across components via \emph{multi-part  cross-attention}.
Unlike standard cross-attention that treats all tokens uniformly,
multi-part interaction is explicitly structured at the part level,
preserving clear part identities while allowing global coordination across components.

Formally, let $\mathbf{X}^k \in \mathbb{R}^{N_k \times d}$ denote the latent tokens of part $k$.
Multi-part interaction is applied after intra-part self-attention
as an explicit inter-part information exchange step.
For each part, contextual information from other components is incorporated by attending to
their latent tokens:
\begin{equation}
\mathbf{X}^{k}_{\text{out}}
=
\mathbf{X}^{k}
+
\mathrm{Attn}\!\left(
\mathbf{Q} = \mathbf{X}^{k},\;
\mathbf{K} = [\mathbf{X}^{j}]_{j \neq k},\;
\mathbf{V} = [\mathbf{X}^{j}]_{j \neq k}
\right),
\label{eq:multipart_attn}
\end{equation}
where $[\cdot]$ denotes concatenation along the token dimension.
The residual formulation preserves each part’s internal representation
while augmenting it with structural context from other components,
enabling coordinated reasoning over relative placement and inter-part relationships
without collapsing parts into a shared representation.

By integrating multi-part latent interaction across all stages of generation,
Seg3DParts produces coherent multi-part assemblies directly in a shared canonical space,
eliminating the need for post-hoc alignment.
Implementation details of the interaction blocks are provided in the appendix.

\subsection{Canonical-Space Part Decoding}
\label{sec:decode}

A key property of Seg3DParts is that all generated part meshes are decoded directly into a shared canonical space, without any post-hoc alignment such as translation, scaling, or optimization.

This is achieved by consistently preserving part-level spatial alignment throughout data preparation and model training.
Parts are obtained by decomposing complete objects while retaining their original scale and placement, and all stages of encoding, generation, and decoding operate directly on these aligned representations.

As a result, at inference time, independently decoded part meshes are already correctly positioned with respect to one another.
This design eliminates the need for explicit assembly and enables stable multi-part generation under varying decomposition granularity and partial observations.

\subsection{Optimization Loss}
\label{sec:loss}

Seg3DParts is trained using a combination of reconstruction and generative objectives across the two stages of the framework.

\noindent\textbf{VAE Training.} Both stage-1 and stage-2 VAEs are trained with a standard variational objective that combines reconstruction losses with KL regularization to encourage a smooth latent distribution.

In the sparse structure stage, the VAE focuses on capturing coarse part-level occupancy, providing a structural prior for downstream generation.
In the structured latent stage, the VAE emphasizes accurate surface geometry reconstruction conditioned on the generated sparse structure.

For stable supervision, decoded representations are normalized before computing reconstruction losses.
Detailed loss formulations and training configurations are provided in the appendix.

\noindent\textbf{DiT Training.}
The rectified-flow DiTs in both stages are trained using a flow matching objective.
Specifically, the DiTs learn to predict velocity fields that transform noise samples into target latent representations under the rectified-flow formulation.

%% file: sec/5_experiments.tex
\section{Experiments}

\subsection{Dataset}

Existing 3D datasets either focus on whole-object reconstruction without explicit part structure, or provide part annotations that are noisy, weakly aligned, and unsuitable for training generative models.
To address this gap, we construct \textbf{PartObjectNet}, a large-scale dataset of high-quality part-separated 3D objects designed specifically for part-level 3D generation.

PartObjectNet is built by integrating data from Objaverse~\cite{objaverse}, Texverse~\cite{zhang2025texverse}, and PartNet~\cite{partnet}.
We first automatically select objects with explicit part-level decompositions, retaining only instances with 2--15 parts to balance compositional diversity and structural complexity.
We further enforce semantic and geometric quality by removing low-quality cases, including white-mesh placeholders, noisy scanned results, large scene-level assets, and objects with unreasonable or semantically inconsistent part decompositions.

After automated filtering followed by manual curation, the final dataset contains approximately 200K high-quality objects spanning diverse categories and decomposition styles.
Each object is represented as a set of part meshes that preserve their original relative scale and spatial placement, providing spatially aligned part-level supervision suitable for learning inter-part relationships and controllable generation.
For evaluation, we randomly hold out 500 objects from PartObjectNet as a test set, with no overlap with the training data.
In addition, to assess generalization beyond the training distribution, we further evaluate our method on an external benchmark, PartObjaverse-Tiny~\cite{SAMPart3D}, which contains objects independent of our training data.
\vspace{-0.8em}
\subsection{Implementation Details}
Seg3DParts is trained following the two-stage pipeline of TRELLIS, with both stages extended to support part-level generation.
In both the sparse structure stage and the structured latents stage, multi-Part Latent Interaction layers are inserted into the VAEs and diffusion transformers to enable cross-part information exchange.
For the structured latents stage, we replace the voxel-feature projection used in TRELLIS with a mesh-based latent encoder similar to TripoSF~\cite{he2025sparseflex}, which improves geometric expressiveness and benefits the reconstruction of occluded regions. Both VAEs are trained using 8 NVIDIA A800 GPUs for approximately two days.
The diffusion transformers in the two stages are trained using rectified-flow objectives on 16 NVIDIA A800 GPUs for approximately one weeks.

\subsection{Evaluation Protocol}

We evaluate Seg3DParts on 3D geometric quality at both the global object level and the part level.
All metrics are computed on the held-out test set of 500 objects and PartObjaverse-Tiny.

\noindent\textbf{Global-Level Geometry Evaluation.}
To assess overall geometric fidelity, we concatenate all generated part meshes into a single mesh and compare it with the ground-truth object.
Generated and ground-truth meshes are normalized into a common $[-1, 1]^3$ space.
We report the Chamfer Distance (CD) and F-Score (FS) at a threshold of $0.1$, computed by uniformly sampling 16K points from each mesh surface.
Lower CD and higher FS indicate better global geometric accuracy.

\noindent\textbf{Part-Level Geometry Evaluation.}
To evaluate individual component quality, each generated part mesh is compared with its corresponding ground-truth part.
We report per-part CD and FS using the same evaluation protocol as the global-level metrics.
In addition, we measure volumetric consistency using Intersection over Union (IoU), where both generated and ground-truth parts are voxelized into a $64 \times 64 \times 64$ grid within the shared canonical space.

\noindent\textbf{Part Overlap Evaluation.}
To assess part disentanglement and spatial separation, we compute the average pairwise IoU between generated parts using the same $64^3$ voxelization.
Lower overlap indicates reduced interpenetration and better geometric decoupling between components.

\begin{figure*}[t]
    \centering
    \includegraphics[width=1.0\linewidth]{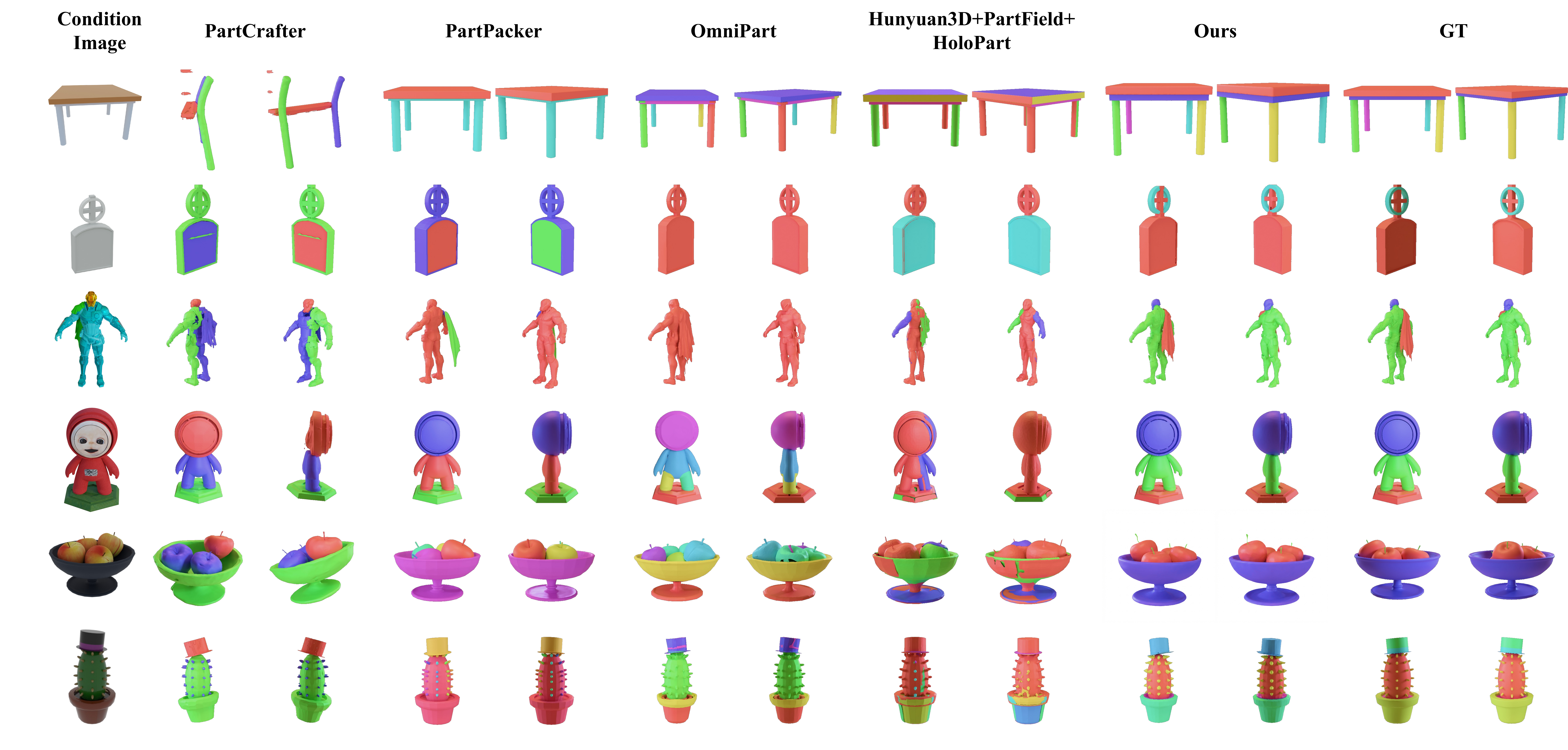}
    \caption{\textbf{Qualitative comparison of part-level 3D generation.}
We compare Seg3DParts with decomposition-based pipelines and part-structured generative models. GT part segmentations are provided as input to Seg3DParts and OmniPart.
Different colors indicate different semantic parts.}
    \vspace{-2em}
    \label{fig:baseline}
\end{figure*}

\subsection{Comparison with State-of-the-Art Methods}

We conduct a comprehensive comparison with state-of-the-art methods following two representative pipelines for part-level 3D generation and reconstruction.

\noindent\textbf{Decomposition-Based Part Generation Pipelines.}
This class of methods follows a sequential strategy that first obtains a complete object mesh and then decomposes it into parts for further processing.
We adopt a representative pipeline combining \textbf{PartField}~\cite{liu2025partfield} and \textbf{HoloPart}~\cite{holopart}.
Specifically, PartField is used to perform surface-based part segmentation on a reconstructed whole-object mesh, while HoloPart completes the geometry of each segmented part.
To obtain the initial whole-object mesh, we employ the state-of-the-art single-view reconstruction model \textbf{Hunyuan3D~2.1}~\cite{zhao2025hunyuan3d}.
This pipeline reflects a commonly used practice where part-level generation is achieved by post-hoc decomposition and completion based on a reconstructed full shape.

\noindent\textbf{Part-Structured Generative Models.}
We further compare against recent generative models that explicitly incorporate part structure into the generative process.
\textbf{OmniPart}~\cite{ominipart} is built entirely upon the TRELLIS framework and performs part-level generation by autoregressively predicting part bounding boxes in the sparse voxel space produced by Stage-1, followed by fine-tuning Stage-2 to synthesize geometry for each predicted part.
\textbf{PartCrafter}~\cite{partcrafter} builds upon a pretrained whole-object generative model and extends it to part-aware generation by introducing explicit part tokens and structured attention.
It alternates local and global attention across DiT layers, enabling parts to be generated jointly while modeling their interactions. \textbf{PartPacker}~\cite{partpacker} is an end-to-end part-level 3D generation framework that utilizes a dual-volume packing strategy. It organizes an arbitrary number of parts into two complementary volumetric latent spaces based on a bipartite contraction formulation. This strategy maintains geometric separation between contacting parts while promoting inter-part consistency through a fixed-length latent representation.

\begin{minipage}{0.45\textwidth}
    \noindent\textbf{Quantitative Results.} We quantitatively evaluate Seg3DParts against state-of-the-art baselines in terms of global geometry fidelity, inter-part overlap, and part-level geometric accuracy.
    As shown in Table~\ref{tab:global} and Table~\ref{tab:part}, Seg3DParts achieves the best overall performance, obtaining the highest F-Score and
\end{minipage}
\hfill
\begin{minipage}{0.5\textwidth}
    {
    \setcounter{table}{1}
    \captionof{table}{
    Part-level geometry evaluation on PartObjectNet and PartObjaverse-Tiny. Only methods with explicit part correspondence are reported.
    }
    \label{tab:part}
    \centering
    \resizebox{1.0\linewidth}{!}{
    \begin{tabular}{c|ccc|ccc}
    \hline
    \multirow{2}{*}{Method}
    & \multicolumn{3}{c|}{PartObjectNet}
    & \multicolumn{3}{c}{PartObjaverse-Tiny} \\
    \cline{2-7}
    & FS@0.1$\uparrow$ & CD$\downarrow$ & IoU$\uparrow$
    & FS@0.1$\uparrow$ & CD$\downarrow$ & IoU$\uparrow$ \\
    \hline
    
    OmniPart & {0.565} & {0.417} & {0.551}
             & {0.455} & {0.418} & {0.429} \\
    
    \textbf{Ours} 
             & \best{0.774} & \best{0.192} & \best{0.781}
             & \best{0.639} & \best{0.310} & \best{0.702} \\
    
    \hline
    \end{tabular}
    }
    }
\end{minipage}
lowest Chamfer Distance among all evaluated methods, indicating superior global geometric fidelity.
In addition, our method produces a substantially lower part-overlap IoU, suggesting that individual components are generated with clearer spatial separation and less interpenetration.
These improvements stem from generating multiple parts directly in a shared canonical space, where part geometry and spatial relationships are modeled jointly, while segmentation-aware conditioning explicitly specifies which component is being generated, and multi-part latent interaction allows each part to adapt its geometry in response to other components.

We further evaluates part-level geometric accuracy for methods that preserve explicit part correspondence.
Among existing approaches, only OmniPart supports conditioning on a given part segmentation and maintains a stable correspondence between generated parts and ground-truth components, making it suitable for fair part-level comparison.
Seg3DParts significantly outperforms OmniPart across all metrics, with large gains in both F-Score and Chamfer Distance.
These results indicate that Seg3DParts not only improves global shape quality, but also produces more accurate and well-aligned individual components, validating its effectiveness for controllable part-level 3D generation.


{
\setcounter{table}{0}
\begin{table}[t]
\caption{
Global geometry (FS, CD) and part-overlap (IoU) evaluation on both PartObjectNet and PartObjaverse-Tiny.}
\label{tab:global}
\centering
\resizebox{\linewidth}{!}{
\begin{tabular}{c|ccc|ccc}
\hline
\multirow{2}{*}{Method} 
& \multicolumn{3}{c|}{PartObjectNet} 
& \multicolumn{3}{c}{PartObjaverse-Tiny} \\
\cline{2-7}
& FS@0.1$\uparrow$ & CD$\downarrow$ & IoU$\downarrow$
& FS@0.1$\uparrow$ & CD$\downarrow$ & IoU$\downarrow$ \\
\hline

PartCrafter                     & 0.698 & 0.248 & 0.050 
                                & 0.731 & 0.182 & 0.049 \\

PartPacker                      & \third{0.876} & \third{0.115} & \third{0.033} 
                                & \second{0.802} & \second{0.138} & \third{0.033} \\

OmniPart                        & \second{0.885} & \second{0.108} & 0.057 
                                & {0.763} & 0.168 & 0.041 \\

Hunyuan3D2.1 + PartField        & 0.860 & 0.120 & \second{0.031} 
                                & 0.793 & 0.144 & \second{0.028} \\

Hunyuan3D2.1 + PartField + HoloPart 
                                & 0.858 & 0.121 & 0.067 
                                & \third{0.796} & \third{0.143} & 0.041 \\ \hline

\textbf{Ours}                   & \best{0.917} & \best{0.085} & \best{0.012} 
                                & \best{0.810} & \best{0.128} & \best{0.018} \\

\hline
\end{tabular}
}
\vspace{-2em}
\end{table}
}





\noindent\textbf{Qualitative Results.} Figure~\ref{fig:baseline} presents qualitative comparisons of part-level 3D generation from a single image.
PartCrafter is able to generate diverse component geometries, but often struggles to maintain consistent semantic part separation, leading to fragmented or ambiguous part boundaries.
PartPacker and decomposition-based pipelines Hunyuan3D + PartField + HoloPart generally produce plausible overall shapes; however, their part-level controllability is unstable, with inconsistent part identities and noticeable variations in relative placement across instances.
Although OmniPart explicitly models part structure by conditioning on a whole-object segmentation map, it frequently exhibits incomplete or merged parts, indicating challenges in robust part separation and identity preservation.
In contrast, Seg3DParts consistently generates complete, well-separated part meshes with accurate alignment and coherent spatial relationships across diverse object categories.
These qualitative results demonstrate the effectiveness of segmentation-aware conditioning and explicit cross-part interaction for reliable part-level 3D generation.

\vspace{-1.3em}
\subsection{Ablation Study}
We conduct ablations to evaluate each component of our framework.
For efficiency, we train all variants on 1K randomly sampled objects for 5K steps under a consistent setup.
Quantitative and qualitative results are shown in Table~\ref{tab:abl} and Fig.~\ref{fig:ablation}, respectively.

\noindent\textbf{Effect of Multi-Part Latent Interaction.}
We first remove the proposed multi-part latent interaction, replacing all multi-part blocks with single-part blocks while keeping all other components unchanged.
As shown in Table~\ref{tab:abl} and Fig.~\ref{fig:ablation}, this variant exhibits a consistent degradation across all metrics.
While the drop in global geometry quality is moderate, the generated parts tend to intersect or overlap in 3D space, as visualized in Fig.~\ref{fig:ablation}.
This indicates that without explicit cross-part interaction, parts are generated more independently, leading to weaker structural coordination and increased interpenetration.

\noindent\textbf{Effect of Segmentation-Aware Part Conditioning.}
We further replace part-wise segmentation conditioning with a single whole-object segmentation map.
Although global shape quality remains comparable, Fig.~\ref{fig:ablation} shows that part boundaries become ambiguous and several components are incorrectly shaped or misplaced.
This results in noticeably degraded part-level fidelity and overlap metrics, highlighting the importance of localized, part-specific conditioning for accurate part synthesis.

\begin{minipage}{0.48\textwidth}
    \noindent\textbf{Controllability via Segmentation Input.}
    Beyond improving part fidelity, segmentation-aware conditioning also enables explicit control over part placement.
    As shown in Fig.~\ref{fig:ablation_diff_seg}, we fix the input image and vary only the part segmentation.
    The generated results faithfully follow the provided segmentation, producing distinct and consistent part layouts for the same object.
    This demonstrates that Seg3DParts allows precise, segmentation-driven control over part decomposition and spatial configuration, rather than relying on a fixed or implicit part structure.
\end{minipage}
\hfill
\begin{minipage}{0.5\textwidth}
    \centering
    \includegraphics[width=\textwidth]{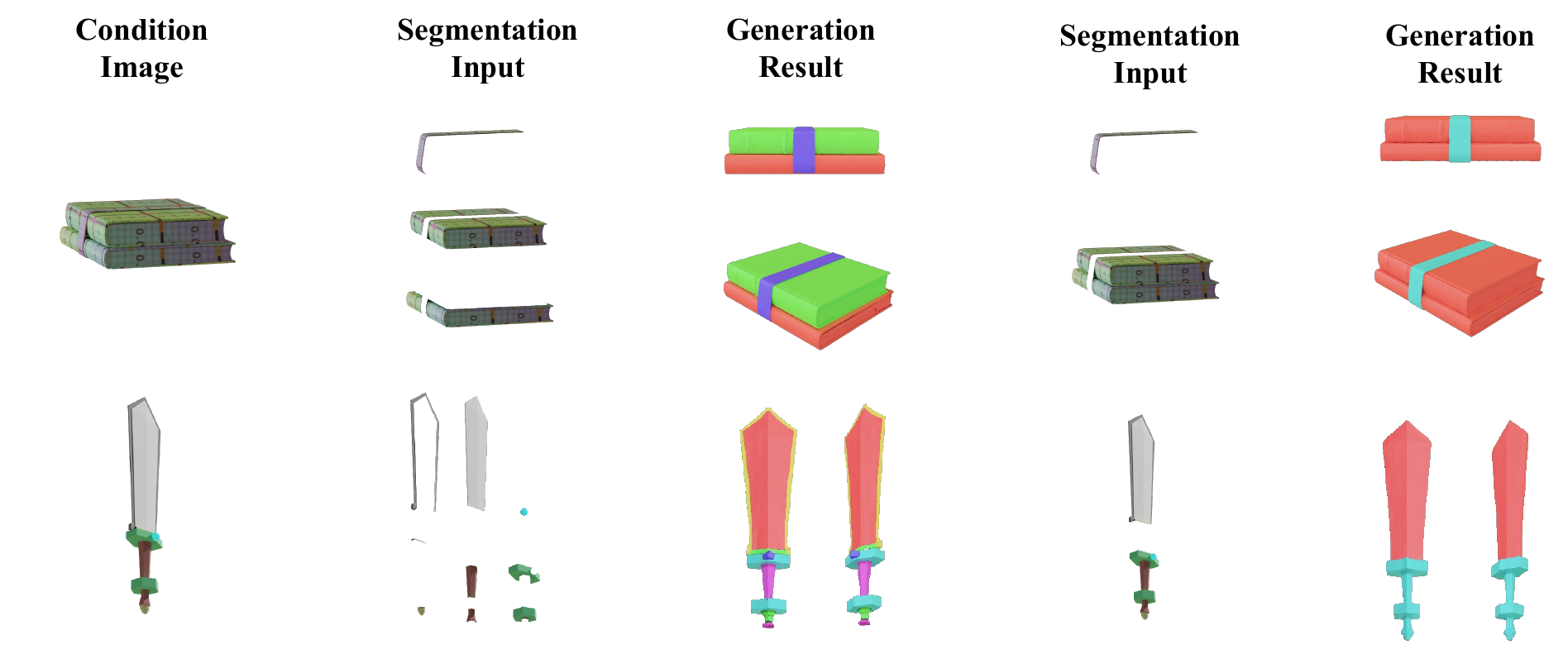}
    \addtocounter{figure}{1}
    \captionof{figure}{\textbf{Segmentation-driven part control.}
Different segmentation inputs on the same image produce distinct and accurate part layouts.}
    \label{fig:ablation_diff_seg}
\end{minipage}

\begin{figure}
    \centering
    \includegraphics[width=0.8\linewidth]{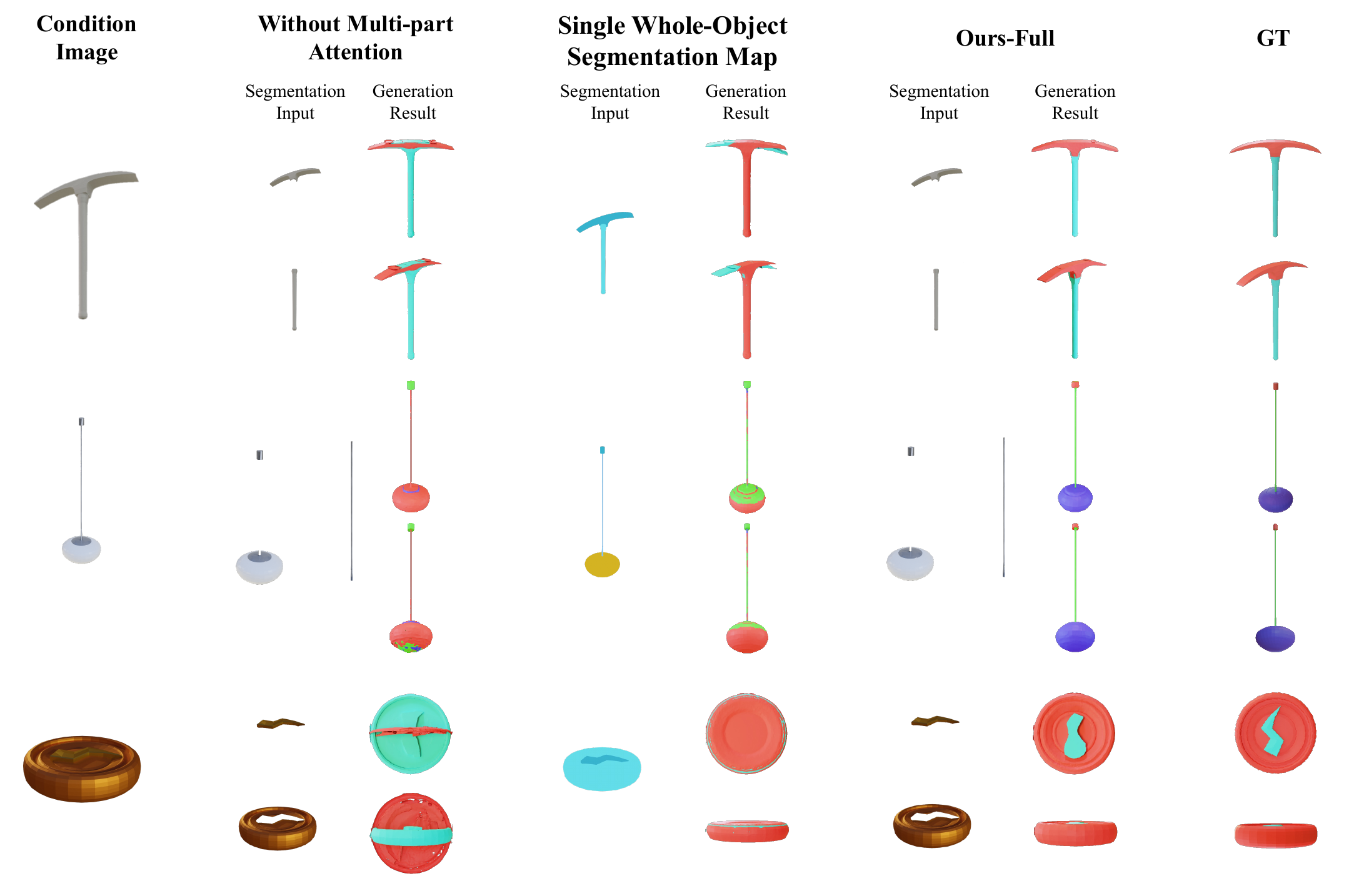}
    \addtocounter{figure}{-2}
    \caption{\textbf{Ablation study on part-level generation.} Removing multi-part attention leads to interpenetrating and incoherent parts, while replacing part-wise segmentation with a single object mask causes unstable decomposition. The full model produces well-separated, structurally coherent parts that closely match the ground truth.
    }
    \vspace{-1.5em}
    \label{fig:ablation}
\end{figure}

\begin{table}[]
\setcounter{table}{2}
\caption{Ablation study of multi-part latent interaction and segmentation-aware part conditioning.
  Metrics are reported for global geometry, part overlap, and part-level accuracy.}
\label{tab:abl}
\centering
\centering
\resizebox{\linewidth}{!}{
\begin{tabular}{c|cc|c|ccc}
\hline
\multirow{2}{*}{Method} & \multicolumn{2}{c|}{Global-Level} & Part-Overlap & \multicolumn{3}{c}{Part-Level} \\
\cline{2-3}\cline{4-4}\cline{5-7}

                  & FS@0.1$\uparrow$ & CD$\downarrow$ & IoU$\downarrow$ & FS@0.1$\uparrow$ & CD$\downarrow$ & IoU$\uparrow$ \\ \hline

Full Model                & \best{0.856} & \best{0.129} & \best{0.054} & \best{0.631} & \best{0.467} & \best{0.649} \\

w/o Multi-Part Interaction       & 0.838 & 0.138 & 0.103 & 0.531 & 0.534 & 0.434 \\

Single Whole-Object Conditioning & 0.840 & 0.137 & 0.126 & 0.488 & 0.589 & 0.444 \\ \hline
\end{tabular}
}
\vspace{-1.5em}
\end{table}


%% file: sec/6_conclusion.tex
\section{Conclusion}

We presented Seg3DParts, a segmentation-based framework for controllable part-level 3D generation, which explicitly leverages 2D semantic segmentation to guide structured 3D part synthesis from a single RGB image.
By combining explicit part grounding with joint multi-part reasoning, our method directly produces coherent and well-aligned part meshes without post-hoc alignment.
Experiments demonstrate that Seg3DParts achieves stronger geometric fidelity and part-level controllability than existing baselines.

%% file: sec/7_appendix.tex
\section*{Appendix}
\setcounter{figure}{0}

\section{Implementation and Architectural Details}

This appendix provides detailed descriptions of the architectural design, training configuration, and inference procedure of Seg3DParts.
These details complement the main paper and are omitted from the core sections for clarity.

\section{Part-Aware Variational Autoencoders}
\label{app:vae}

Seg3DParts adopts a two-stage variational autoencoding scheme for modeling part-level
3D geometry, following the structured latent generation paradigm of TRELLIS~\cite{xiang2025structured}.
Both stages operate on part-separated representations, where each semantic component
is encoded and decoded independently while sharing network parameters.
All VAEs are trained with standard variational objectives that combine reconstruction
losses with KL divergence regularization.
Decoded representations are normalized for stable supervision, while the canonical
alignment of parts is preserved throughout training.

\subsection{Stage-1 Part-Aware Sparse Structure VAE}
\label{app:vae_stage1}

The stage-1 VAE models coarse part-level structure using a sparse voxel occupancy
representation at $64^3$ resolution.
To support part-level generation, each semantic part is treated as an independent
element along the batch dimension.
This design allows all parts to be processed in parallel with shared encoder--decoder
weights, while preserving explicit separation between components.

\noindent\textbf{Architecture.}
The encoder maps each part’s sparse voxel grid into a compact latent representation
using a hierarchical 3D convolutional backbone.
For each part, a Gaussian posterior is inferred by predicting the mean and variance
of the latent distribution.
The decoder mirrors the encoder with symmetric upsampling layers and reconstructs
binary voxel occupancy for each part at the original resolution.

\noindent\textbf{Multi-Part Interaction.}
To capture global structural relationships among components, explicit cross-part
interaction is introduced at the latent bottleneck.
At this stage, part-wise latent feature grids exchange information via attention
along the part dimension, allowing each component to reason about the presence,
relative placement, and spatial extent of other parts while maintaining distinct
identities.
Multi-part interaction is restricted to the bottleneck, where representations are
compact and semantically meaningful.
Higher-resolution encoder and decoder stages operate independently on each part to
preserve local geometric detail.
The interaction is applied as a residual refinement and reduces to standard
self-attention when only a single part is present.

\noindent\textbf{Training Objective.}
The stage-1 VAE is trained to reconstruct sparse voxel occupancy using a Dice loss,
which is well suited for supervising highly sparse structures.
A KL divergence term regularizes the latent distribution.
The training objective is
\begin{equation}
\mathcal{L}_{\text{stage1}}
=
\mathcal{L}_{\text{dice}}
+
\lambda_{\text{KL}}^{\text{ss}} \,
\mathcal{L}_{\text{KL}},
\end{equation}
where the KL weight is set to
$\lambda_{\text{KL}}^{\text{ss}} = 10^{-3}$.
Decoded voxel grids are normalized before loss computation for numerical stability,
without affecting the canonical alignment of part representations.

\subsection{Stage-2 Part-Aware Structured Latent VAE}
\label{app:vae_stage2}

The stage-2 VAE models fine-grained surface geometry for each part.
It adopts a mesh-based structured latent encoding strategy inspired by
TripoSF~\cite{he2025sparseflex}, directly encoding surface geometry into sparse
latent tokens.
Compared to voxel-feature projection, this design provides stronger geometric
expressiveness and improved robustness under partial or occluded observations.

\noindent\textbf{Architecture.}
For each part, surface geometry is encoded into sparse structured latent tokens,
which are processed by a sparse transformer encoder--decoder.
Both the encoder and decoder consist of $12$ transformer blocks and operate on
part-separated latent representations, preserving explicit part identities
throughout the network.
The encoder predicts a Gaussian posterior for each part’s structured latent code.

\noindent\textbf{Multi-Part Latent Interaction.}
To promote geometric coherence across components, multi-part latent interaction
is integrated symmetrically into both the encoder and decoder.
A subset of transformer blocks is replaced with multi-part interaction blocks,
in which latent tokens of each part attend to those of other parts via cross-part
attention, while the remaining blocks process parts independently.
This periodic interaction enables global coordination among parts without
collapsing them into a shared representation, and reduces to standard
self-attention in the single-part case.

\noindent\textbf{Mesh Decoding.}
The decoder progressively upsamples sparse latent features and reconstructs
explicit surface geometry for each part using a FlexiCubes-based mesh extraction
module.
FlexiCubes \cite{shen2023flexicubes} provides a differentiable and topology-adaptive surface representation,
allowing high-quality mesh reconstruction while maintaining training stability.
All part meshes are decoded directly into a shared canonical space, enabling
coherent multi-part assemblies without post-hoc alignment.

\noindent\textbf{Training Objective.}
The stage-2 part-aware structured latent VAE focuses exclusively on geometric reconstruction.
Its training objective combines multiple geometry-aware supervision signals together with
KL regularization on the structured latent distribution:
\begin{equation}
\begin{aligned}
\mathcal{L}_{\text{VAE}}^{\text{sl}}
=
&\;
\mathcal{L}_{\text{mask}}
+ \lambda_{\text{depth}} \, \mathcal{L}_{\text{depth}}
+ \lambda_{\text{tsdf}} \, \mathcal{L}_{\text{tsdf}} \\
&+ \mathcal{L}_{\text{normal}}^{\text{perceptual}}
+ \lambda_{\text{KL}}^{\text{sl}} \, \mathcal{L}_{\text{KL}} ,
\end{aligned}
\label{eq:stage2_vae_loss}
\end{equation}
where $\mathcal{L}_{\text{mask}}$ supervises the rendered silhouette,
$\mathcal{L}_{\text{depth}}$ is computed using a Smooth-$\ell_1$ loss on rendered depth maps,
and $\mathcal{L}_{\text{tsdf}}$ supervises the truncated signed distance field.

The perceptual normal loss is defined as
\begin{equation}
\mathcal{L}_{\text{normal}}^{\text{perceptual}}
=
\mathcal{L}_{\text{normal}}^{\ell_1}
+ \lambda_{\text{ssim}} \, \mathcal{L}_{\text{ssim}}^{\text{normal}}
+ \lambda_{\text{lpips}} \, \mathcal{L}_{\text{lpips}}^{\text{normal}},
\end{equation}
which enforces both low-level and perceptual consistency on rendered surface normals
using a combination of $\ell_1$, SSIM~\cite{ssim}, and LPIPS~\cite{lpips} losses.

The loss weights are set to
$\lambda_{\text{depth}} = 10.0$,
$\lambda_{\text{tsdf}} = 0.01$,
$\lambda_{\text{ssim}} = 0.2$,
$\lambda_{\text{lpips}} = 0.2$,
and $\lambda_{\text{KL}}^{\text{sl}} = 10^{-6}$.
Decoded geometry is normalized to $[-1,1]$ before loss computation for numerical stability,
without altering the canonical alignment of part representations.

\begin{figure}
    \centering
    \includegraphics[
  width=0.8\textwidth,
]{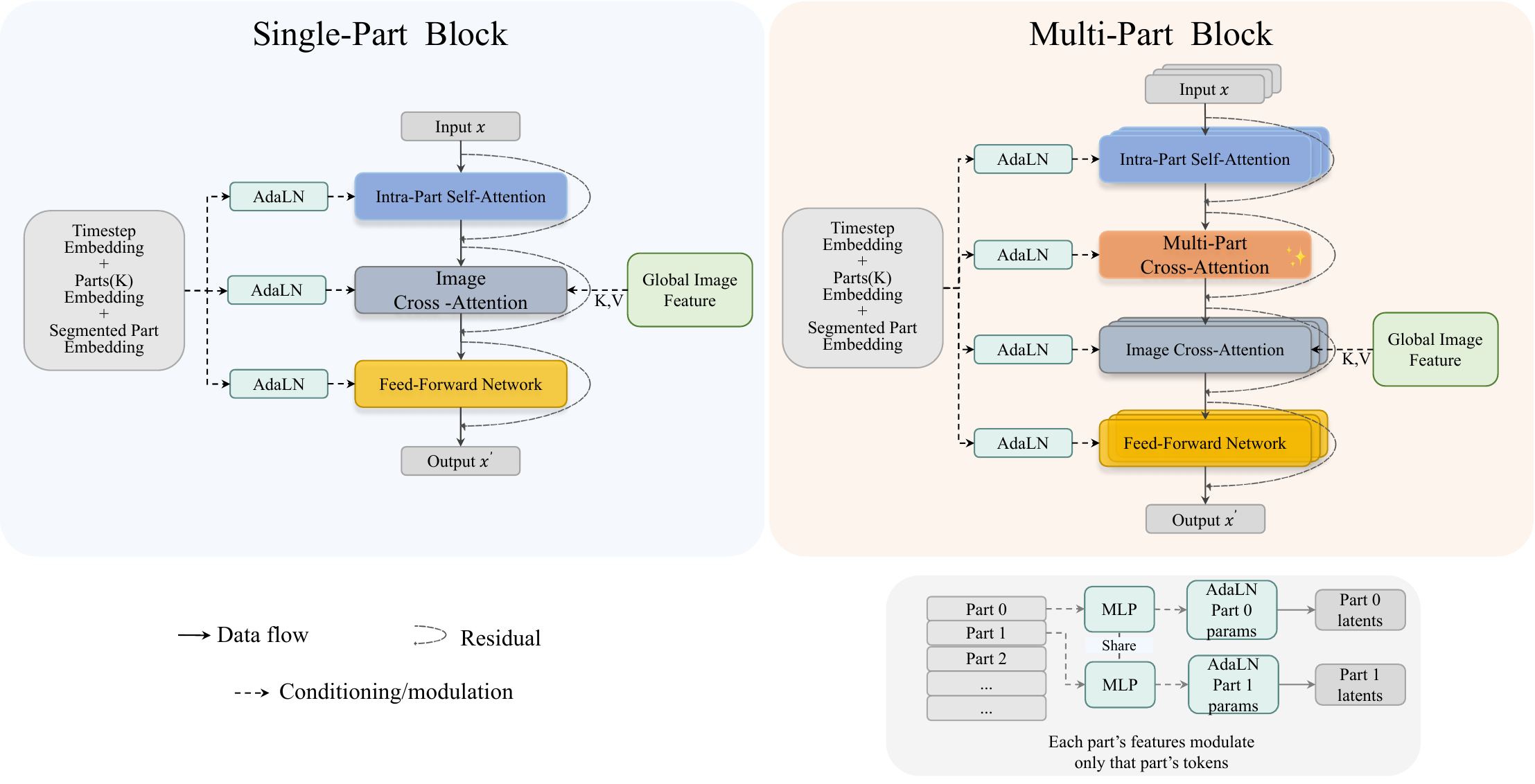}\caption{{\textbf{Single-part and multi-part transformer block in Seg3DParts.}
Single-part block model each semantic part independently using intra-part self-attention,
image cross-attention, and feed-forward layers with AdaLN conditioning.
Multi-part block insert an additional cross-part attention layer to enable information
exchange across parts while preserving explicit part identities.
AdaLN modulation is applied in a part-specific manner, so each part’s conditioning affects
only its own latent tokens.}}
    \label{fig:dit_block}
\end{figure}

\section{Transformer Architecture Details}
\label{app:dit}
The rectified-flow Diffusion Transformers (DiTs) used in both stages of Seg3DParts
share a unified backbone architecture.
Each DiT consists of $24$ transformer blocks with a hidden dimension of $1024$
and $16$ attention heads.
The two stages differ primarily in their input resolution and data representation:
part-aware sparse structure DiT (Stage~1) operates on voxel grids at a resolution of $16^3$ to model coarse
part structure, while part-aware structure latent DiT (Stage~2) processes sparse voxel features embedded in a
$64^3$ coordinate space to capture fine-grained surface geometry.
In both stages, part-level conditioning and cross-part coordination are integrated
into the transformer through a mixture of \emph{single-part blocks} and \emph{multi-part blocks},
as detailed below.

\subsection{Single-Part and Multi-Part Transformer Blocks}
\label{app:dit_blocks}
\noindent\textbf{Single-Part Blocks.}
As illustrated in Fig.~\ref{fig:dit_block}, single-part blocks operate independently on each semantic part.
Each block consists of intra-part self-attention, image-conditioned cross-attention,
and a feed-forward network, with residual connections throughout.
Self-attention is confined to tokens within the same part, allowing local geometric
structure to be modeled without interference from other components.
These blocks constitute the majority of the network and primarily focus on refining
per-part geometric details.

\noindent\textbf{Multi-Part Blocks.}
As illustrated in Fig.~\ref{fig:dit_block}, multi-part blocks extend the single-part design by inserting an additional
\emph{multi-part attention} layer immediately after intra-part self-attention.
This layer enables information exchange across different parts, allowing the model
to coordinate relative placement, scale, and structural compatibility among components.
The remaining layers, including image-conditioned cross-attention and the feed-forward network,
retain the same structure as in single-part blocks.

Multi-part blocks are sparsely interleaved throughout the transformer according to a fixed schedule,
providing periodic global coordination while preserving part-level specialization.
When only a single part is present, the multi-part attention naturally reduces to
standard self-attention.

\subsection{Conditioning with AdaLN}
\label{app:adln}

As illustrated in Fig.~\ref{fig:dit_block}, both single-part and multi-part blocks
use AdaLN for conditioning.
The conditioning signal is formed by combining the timestep embedding, a parts-count embedding,
and a per-part embedding extracted from the segmented image region.
These per-part embeddings are processed by a shared MLP to produce AdaLN parameters
(e.g., scale, shift, and gating) for each part, and are then applied to the corresponding
part tokens via indexed broadcasting, so that the conditioning of part $k$ affects only
the tokens belonging to part $k$.

\section{Inference and Sampling Procedure}
\label{app:sampling}

At inference time, Seg3DParts follows the same two-stage generation pipeline as training.
Both sparse structure and structured latent DiTs are sampled using rectified-flow integration with sampling steps 50 and 30 respectively.
Classifier-free guidance (CFG) is applied to image conditioning with a 3.0 guidance scale across all experiments.
No test-time optimization or post-processing is used.
All parts are decoded independently and naturally aligned in a shared canonical space.

\section{Dataset Construction Details}
\label{app:dataset}

PartObjectNet is constructed by aggregating part-decomposed 3D objects from
Objaverse, Texverse, and PartNet.
We retain only objects with 2--15 semantic parts, which covers common composite
structures while avoiding trivial or excessively fragmented cases.
All candidates are further filtered through automatic checks and manual curation
to remove low-quality meshes, scanned artifacts, scene-level assets, and
semantically inconsistent part decompositions.

For each retained object, all parts preserve their original scale and relative
spatial placement, providing aligned part-level supervision suitable for training
part-aware generative models.
The final dataset contains approximately 200K objects spanning diverse categories
and decomposition styles.


\section{Fully Occluded Parts}
\label{app:occluded}

\begin{figure}
    \centering
    \includegraphics[width=\textwidth]{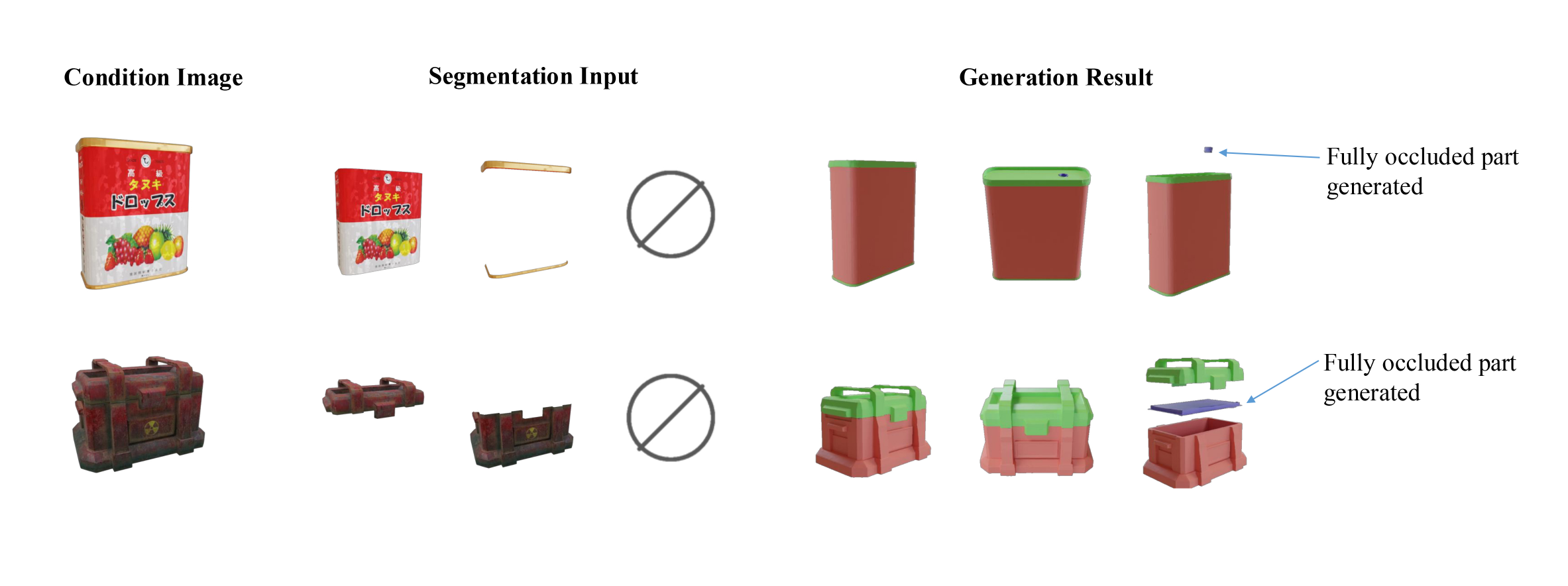}
    \caption{\textbf{Generation of fully occluded parts.} Given a single condition image (left), Seg3DParts generates a complete part decomposition (right) even for parts fully hidden in the input view, which receive an all-black image condition. Arrows indicate the recovered occluded parts: the circular pour-out opening on top of the juice tin (top) and the inner lid of the toolbox (bottom).}
    \label{fig:ablation_occluded}
\end{figure}

Seg3DParts can also generate parts that are fully occluded in the input image.
For a fully occluded part, we supply an all-black image crop as its part image condition.
This is consistent with the training procedure: whenever a part is entirely hidden in the training view, it likewise receives an all-black image condition.
The model therefore learns to interpret a blank image condition as the signal that no visual evidence is available for that component, and relies on cross-part attention to infer its shape and placement from the visible parts (Fig.~\ref{fig:ablation_occluded}).

\section{Limitations}
\label{app:limitations}

\begin{figure}[h]
    \centering
    \includegraphics[width=\textwidth]{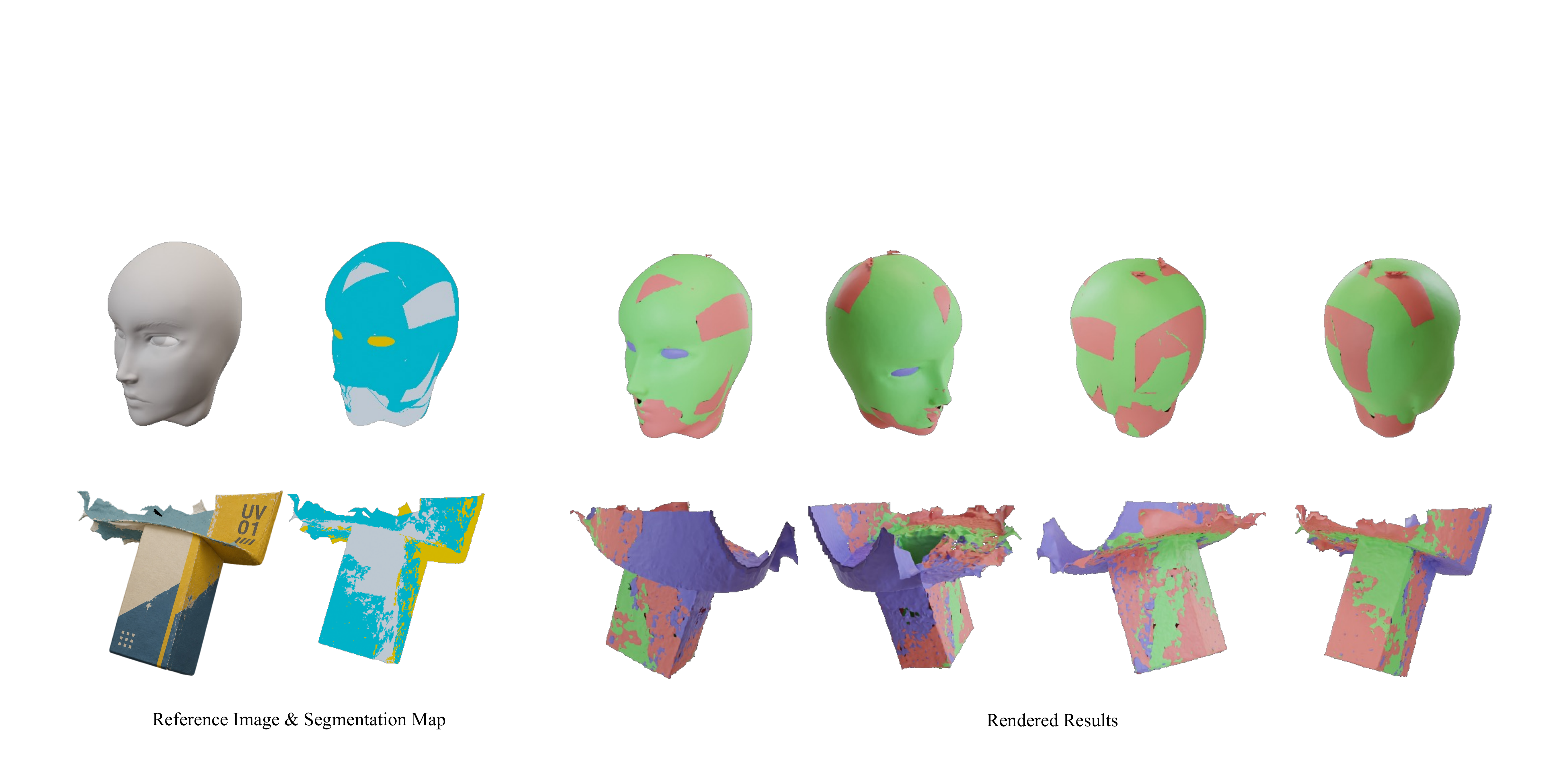}
    \caption{\textbf{Failure cases.} Seg3DParts may produce degraded geometry when the input part segmentation is of low quality or when part boundaries are semantically ambiguous.}
    \label{fig:limitations}
\end{figure}

While Seg3DParts enables flexible and coherent part-level generation, its performance depends on the quality of the input part segmentation.
When the segmentation is of low quality or the part boundaries are semantically ambiguous, the generated geometry may degrade accordingly
(Fig.~\ref{fig:limitations}).

\section{More Results}
\label{app:more_results}

\begin{figure*}[h]
    \centering
    \includegraphics[width=\textwidth]{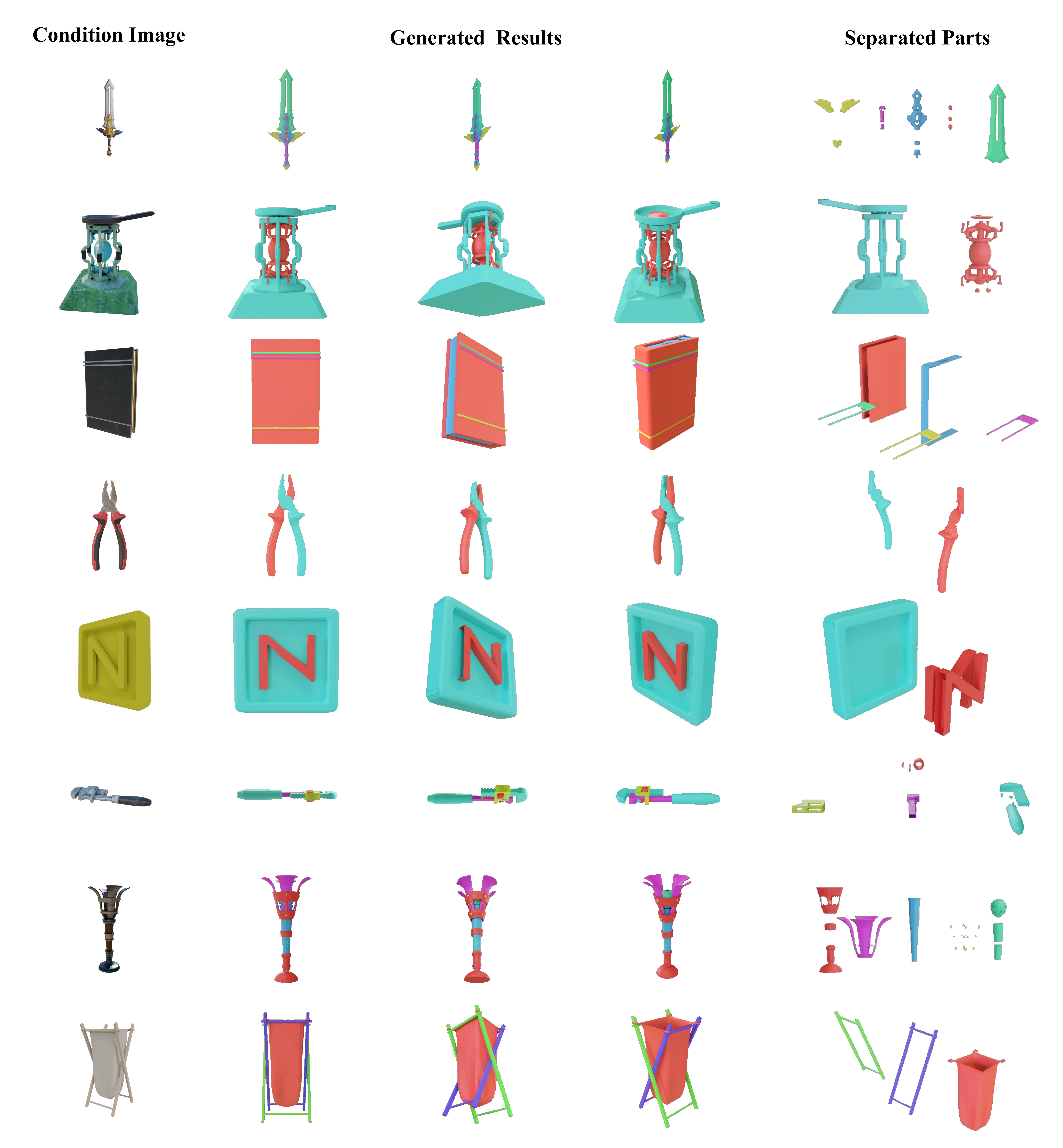}
    \caption{\textbf{Additional qualitative results (Part I)}.}
    \label{fig:more_results_1}
\end{figure*}

\begin{figure*}[h]
    \centering
    \includegraphics[width=\textwidth]{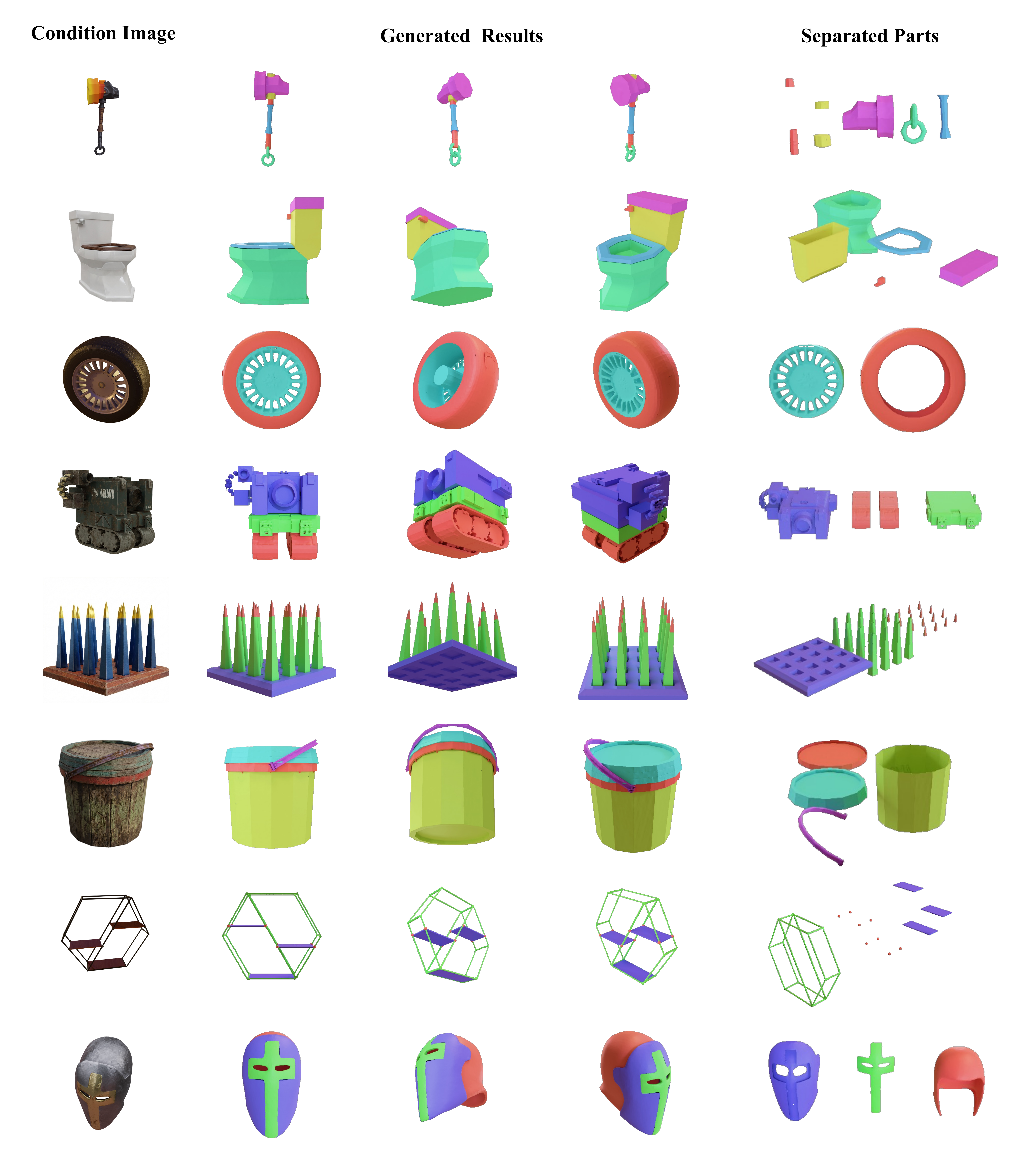}
    \caption{\textbf{Additional qualitative results (Part II).}.}
    \label{fig:more_results_2}
\end{figure*}

Figures~\ref{fig:more_results_1} and~\ref{fig:more_results_2} show additional qualitative results of Seg3DParts across diverse object categories and part decompositions, further demonstrating the generalization and coherence of our method.

